%% file: neurips_2024.tex
\documentclass{article}

\PassOptionsToPackage{numbers, compress}{natbib}

\usepackage[preprint]{neurips_2024}

\usepackage[utf8]{inputenc} 
\usepackage[T1]{fontenc}    
\usepackage{hyperref}       
\usepackage{url}            
\usepackage{booktabs}       
\usepackage{amsfonts}       
\usepackage{nicefrac}       
\usepackage{microtype}      
\usepackage{xcolor}         

\usepackage{comment}
\usepackage{multirow}
\usepackage{makecell}
\usepackage[pdftex]{graphicx}
\usepackage{amsmath,amsfonts,amsthm}
\usepackage{mathtools}
\usepackage{adjustbox}
\usepackage{multicol}
\usepackage{multirow}
\usepackage[linesnumbered,ruled]{algorithm2e}
\usepackage{amsmath}
\usepackage{caption}
\usepackage{subcaption}
\usepackage{listings}
\usepackage{wrapfig}
\usepackage{tablefootnote}
\usepackage{changepage}

\title{GPT-FL: Generative Pre-Trained Model-Assisted Federated Learning}

\author{%
  Tuo Zhang$^{1}$\thanks{The first two authors contribute equally}\footnotemark[1] , Tiantian Feng$^{1}$\footnotemark[1] , Samiul Alam$^{2}$, \\  
  \textbf{Dimitrios Dimitriadis$^{3}$, Sunwoo Lee$^{4}$,} \\ 
  \textbf{Mi Zhang$^{2}$, Shrikanth S. Narayanan$^{1}$, Salman Avestimehr$^{1}$}
  \\
$^1$University of Southern California, $^2$The Ohio State University, $^3$Amazon, $^4$Inha University\\
{\small\texttt{tuozhang@usc.edu, tiantiaf@usc.edu}}\\}

\begin{document}

\maketitle

\begin{abstract}
In this work, we propose \texttt{GPT-FL}, a generative pre-trained model-assisted federated learning (FL) framework. At its core, \texttt{GPT-FL} leverages generative pre-trained models to generate diversified synthetic data. These generated data are used to train a downstream model on the server, which is then fine-tuned with private client data under the standard FL framework. 
We show that \texttt{GPT-FL} consistently outperforms state-of-the-art FL methods in terms of model test accuracy, communication efficiency, and client sampling efficiency. 
Through comprehensive ablation analysis across various data modalities, we discover that the downstream model generated by synthetic data plays a crucial role in controlling the direction of gradient diversity during FL training, which enhances convergence speed and contributes to the notable accuracy boost observed with \texttt{GPT-FL}. Also, regardless of whether the target data falls within or outside the domain of the pre-trained generative model, \texttt{GPT-FL} consistently achieves significant performance gains, surpassing the results obtained by models trained solely with FL or synthetic data.
The code is available at \url{https://github.com/AvestimehrResearchGroup/GPT-FL}.
\end{abstract}

\input{1_intro}

\input{2_related}
\input{3_design}

\input{theory}
\input{4_exp}

\input{5_conclusion}

\bibliographystyle{plain}
\bibliography{main}


\appendix
\input{appendix}

\end{document}

%% file: 1_intro.tex
\section{Introduction} \label{sec:intro}
\vspace{1mm}
Federated learning (FL) is a privacy-preserving machine learning paradigm that allows a collection of clients to collaboratively train a machine learning model without sharing their private data~\cite{Zhang2021FederatedLF}. 
Most existing FL studies such as~\cite{McMahan2016CommunicationEfficientLO, Bonawitz2019TowardsFL} follow the standard FL architecture, where each participating client trains a local model using its own private data and a central server aggregates these locally trained models to update a global model and sends it back to the clients for the next round of training. 
However, although many efforts have been made~\cite{Sahu2018FederatedOI, Karimireddy2019SCAFFOLDSC, Reddi2020AdaptiveFO, Mishchenko2022ProxSkipYL}, the performance of standard FL is still constrained due to the heterogeneity in private data distribution across the clients.

To enhance the performance of FL, recent studies propose to incorporate data collected from public spaces such as the Internet into the FL process~\cite{Lin2020EnsembleDF, Li2021ModelContrastiveFL, Itahara2020DistillationBasedSF, Cho2022HeterogeneousEK}. However, the performance of such public data-based approaches is heavily dependent on the quality of the collected public data. Unfortunately, obtaining the desired public data can be extremely challenging in practice and there is a lack of principled guidance on how to obtain them.
To address the issues of public data-based approaches, FL methods based on synthetic data emerge~\cite{Zhang2022FinetuningGM, Zhu2021DataFreeKD, pi2022dynafed, Wijesinghe2023PSFedGANAE}. 
In~\cite{Zhang2022FinetuningGM, Zhu2021DataFreeKD}, a generative model is trained through knowledge distillation (KD) and the synthetic data are generated from the generative model in an \textit{interleaved} manner \textit{throughout} the federated training iterations. Unfortunately, these approaches are confronted with two limitations: 
(1) since the training of the generative model and the federated training process interleave, the quality of the synthetic data generated by the generative model before it converges can be extremely unstable. Such low-quality synthetic data would in turn jeopardize the federated training process;
(2) given that KD requires clients to report model weights as teachers to transfer knowledge, they are incompatible with secure aggregation protocols~\cite{Bonawitz2017PracticalSA, So2021LightSecAggAL}, which limits their privacy guarantee compared to standard FL.

In this work, we propose \texttt{GPT-FL}, a generative pre-trained model-assisted FL framework that effectively addresses the issues of existing methods. 
The key idea behind \texttt{GPT-FL} is to leverage the knowledge from the generative pre-trained models and to \textit{decouple} synthetic data generation from the federated training process. 
Specifically, \texttt{GPT-FL} prompts the generative pre-trained models to generate diversified synthetic data. These generated data are used to train a downstream model on the server in the centralized manner, which is then fine-tuned with the private client data under the standard FL framework. 
By doing this, the proposed \texttt{GPT-FL} is able to combine the advantages of previous methods while addressing their limitations.

\begin{table*}[t]
\vspace{-11mm}
\caption{Comparison of \texttt{GPT-FL} with existing FL methods.}
\label{comparison-table}
\begin{center}
\resizebox{1.0\linewidth}{!}{
\begin{tabular}{lccccccc}
\toprule  
& \begin{tabular}[c]{@{}c@{}c@{}}\textbf{External Data}\end{tabular} &
\begin{tabular}[c]{@{}c@{}c@{}}\textbf{Limited to} \\ \textbf{Smaller}\\ \textbf{Client Model}\end{tabular} & \begin{tabular}[c]{@{}c@{}c@{}}\textbf{Generate}\\\textbf{Data}\\\textbf{ during FL}\end{tabular}
& \begin{tabular}[c]{@{}c@{}c@{}}\textbf{Data }\\\textbf{Generator}\\\textbf{Location}\end{tabular}
& \begin{tabular}[c]{@{}c@{}c@{}}\textbf{Client Access to}\\\textbf{Public/Generated}\\\textbf{Data}\end{tabular} & 
\begin{tabular}[c]{@{}c@{}c@{}}\textbf{Support }\\\textbf{Data}\\\textbf{Modality}\end{tabular} &
\multicolumn{1}{l}{\begin{tabular}[c]{@{}c@{}c@{}}\textbf{Compatibility }\\\textbf{with Secure}\\\textbf{Aggregation}\end{tabular}}
\\
\cmidrule{1-8}
FedAvg~\cite{McMahan2016CommunicationEfficientLO} & \multirow{4}{*}{No}  & \multirow{4}{*}{No} & \multirow{4}{*}{N/A} & \multirow{4}{*}{N/A} & \multirow{4}{*}{N/A} & \multirow{4}{*}{\textbf{Image/Audio/Text}} &\multirow{4}{*}{Yes} \\
FedOpt~\cite{Reddi2020AdaptiveFO} & & & & & &\\
FedProx~\cite{Sahu2018FederatedOI} & & & & & &\\
SCAFFOLD~\cite{Karimireddy2019SCAFFOLDSC} & & & & & &\\
\cmidrule{1-8}
FedDF~\cite{Lin2020EnsembleDF} & \multirow{3}{*}{\begin{tabular}[c]{@{}c@{}}Public Data \end{tabular}} & \multirow{3}{*}{No} & \multirow{3}{*}{N/A} & \multirow{3}{*}{N/A} & Not Required & \multirow{3}{*}{\textbf{Image/Audio/Text}} &\multirow{3}{*}{No} \\
DS-FL~\cite{Itahara2020DistillationBasedSF} & & & & & Required &  &\\
Fed-ET~\cite{Cho2022HeterogeneousEK} & & & & & Not Required &  &\\
\cmidrule{1-8}
FedGen~\cite{Zhu2021DataFreeKD} & \multirow{4}{*}{\begin{tabular}[c]{@{}c@{}}Generated Data \end{tabular}} & \multirow{3}{*}{Yes} & \multirow{3}{*}{Yes} & Client & Required & Only Image & No\\
FedFTG~\cite{Zhang2022FinetuningGM} & & & & Server & Not Required & Only Image & No\\
DynaFed~\cite{pi2022dynafed} & & & & Server & Not Required & Only Image & Yes\\
\textbf{GPT-FL (ours)} &  & \textbf{No} & \textbf{No} & \textbf{Server} & \textbf{Not Required} & \textbf{Image/Audio/Text} &\textbf{Yes} \\
\bottomrule
\end{tabular}
}
\end{center}
\vspace{-5mm}
\end{table*}

The proposed \texttt{GPT-FL} exhibits multifold merits compared to prior arts (Table~\ref{comparison-table}): 
(1) In contrast to public data-based FL methods, \texttt{GPT-FL} gets rid of the dependency on the availability of the desired public data, offering much more flexibility in its applications. 
(2) Compared to other generative data-based approaches, the leverage of generative pre-trained models and the decoupling between synthetic data generation from the federated training process make the generated synthetic data in \texttt{GPT-FL} not impacted by private data distribution on the clients and the structure of the model to be trained. 
(3) By leveraging the computational resources on the server, \texttt{GPT-FL} provides a much more efficient way to utilize external data by incorporating them into the pre-training of the downstream model, which significantly reduces the communication and computation costs of FL.
(4) The generation of downstream models using synthetic data takes place on the server. As such, it thereby eliminates the need for clients to bear any additional computational burden.
(5) Lastly, as \texttt{GPT-FL} does not alter the standard FL framework, it is fully compatible with secure aggregation protocols as in standard FL methods. More importantly, \texttt{GPT-FL} does not introduce any additional hyper-parameters beyond the standard FL framework. This significantly simplifies the hyper-parameter optimization process, making \texttt{GPT-FL} much more practically useful.

We evaluate the performance of \texttt{GPT-FL} by comparing it against state-of-the-art FL methods under three categories: standard FL methods, public data-based methods, and generated data-based methods on five datasets that cover both image and audio data modalities. 
We highlight five of our findings: 
(1) \texttt{GPT-FL} consistently outperforms state-of-the-art FL methods under both low and high data heterogeneity scenarios with significant advantages in communication and client sampling efficiency.
(2) Under a zero-shot setting, \textit{i.e.} no real-world data is available, the downstream model after centralized training with synthetic images as part of \texttt{GPT-FL} achieves higher performance compared to the global model based on standard FL training with private data. On the contrary, the centralized training with synthetic audio performs worse than FL setups due to the impact of data modality and the quality of the generative pre-trained models.
(3) GPT-FL does not fully rely on generated data. Regardless of whether the target data falls within or outside the domain of the pre-trained generative model, \texttt{GPT-FL} can largely improve model performance beyond relying solely on private data in a standard FL framework.
(4) The downstream model generated by synthetic data controls gradient diversity during FL training, improving convergence speed and leading to significant accuracy gains with \texttt{GPT-FL}.
(5) \texttt{GPT-FL} effectively leverages existing pre-trained downstream models to improve performance in the FL setting, similar to methods under the standard FL framework.

%% file: 2_related.tex
\vspace{-1mm}
\section{Related Work} \label{related}
\vspace{-1mm}

\textbf{Standard Federated Learning.} 
In standard federated learning (FL), clients perform local model training on their private data whereas the central server aggregates these locally trained models to update a global model, which is then sent back for the next round of training. To enhance privacy, Secure Aggregation (SA) protocols~\cite{Bonawitz2017PracticalSA, So2021LightSecAggAL} have been proposed to encrypt each model update and reveal only the sum of the updates to the server. However, the performance of FL is jeopardized by client drift which is caused by the heterogeneity of private data distribution. To tackle this issue, 
previous research proposed various aggregation functions to control the client drift for the global model, such as FedProx~\cite{Sahu2018FederatedOI}, SCAFFOLD~\cite{Karimireddy2019SCAFFOLDSC}, FedOpt~\cite{Reddi2020AdaptiveFO}, and ProxSkip~\cite{Mishchenko2022ProxSkipYL,Malinovsky2022VarianceRP}.


\vspace{0mm}
\noindent
\textbf{FL with Public Data.}
To further mitigate client drift, recent studies propose to utilize  public data (e.g., collected from the internet) in the process of federated training. For example, FedDF, DS-FL, and Fed-ET~\cite{Lin2020EnsembleDF, Itahara2020DistillationBasedSF, Cho2022HeterogeneousEK} leverage public data at the server to aggregate client models through knowledge distillation (KD). 
%
%
%
Mixed FL~\cite{Augenstein2022MixedFL} proposes offloading some intensive computations from clients to the server with public data to reduce client-side load.
%
%
%
%
Some recent studies~\cite{Wang2023CanPL, Hou2023PrivatelyCP} offers to leveraging public datasets and Large Language Models (LLMs) for differentially private on-device FL model training, enhancing the privacy-utility tradeoff through distillation in NLP tasks.
However, utilizing public data for FL has several limitations: the performance of FL heavily relies on the selected public data. Also, it is unclear to which extent should the public data be related to the training data to guarantee effective knowledge distillation, making it challenging to find appropriate public data for every use case~\cite{Stanton2021DoesKD, Alam2022FedRolexMF, Zhang2022FinetuningGM}. 
Moreover, the involvement of KD requires clients to send model weights to the server, which is incompatible with secure aggregation protocols, making them vulnerable to backdoor attacks~\cite{Wang2020AttackOT}. Furthermore, some proposed methods~\cite{Li2021ModelContrastiveFL, Lin2020EnsembleDF} require clients to process the public data. Such requirement adds an extra computational burden to clients.



\vspace{0mm}
\noindent
\textbf{FL with Synthetic Data.}
To address the issues of public data-based approaches, FL methods based on synthetic data have been proposed~\cite{Zhang2022FinetuningGM, Zhu2021DataFreeKD, pi2022dynafed, Wijesinghe2023PSFedGANAE}.
%
In particular, FedGen~\cite{Zhu2021DataFreeKD} and FedFTG~\cite{Zhang2022FinetuningGM} propose to train a lightweight generator on the server using an ensemble of local models in a data-free manner to assist global model training.
%
However, training of the generator relies heavily on the global model, which can lead to poor performance under high data heterogeneity.
%
Additionally, the quality of training the generator is impacted by the structure of the global model~\cite{Kim2022DatasetCV}, making the quality of the synthetic data unstable during training. These approaches are also limited to image-related tasks, restricting their applicability to other data modalities. Specifically, both FedGen and FedFTG rely on training MLP-based or GAN-based lightweight generator networks to ensemble user information in a data-free manner, where the lightweight generator may have limitations in generating high-fidelity data. In addition, the MLP-based model is impractical to model temporal structures to signals such as audio and speech. Finally, some approaches~\cite{Zhang2022FinetuningGM, Zhu2021DataFreeKD} could not support secure aggregation protocols due to the KD-based training, which could compromise the privacy of client data.
As an alternative, DynaFed~\cite{pi2022dynafed} proposes to generate synthetic data via gradient inversion by applying multi-step parameter matching on global model trajectories 
However, using gradient inversion for generating synthetic data could encounter limitations when dealing with high-resolution images~\cite{Huang2021EvaluatingGI}. In addition, this approach could not be directly used for other data modalities such as audio~\cite{Dang2021AMT}. 
%
In this work, we propose \texttt{GPT-FL} as a solution to address these limitations. 

%% file: 3_design.tex
\section{GPT-FL: Generative Pre-Trained Model-Assisted Federated Learning} \label{method}
%
%

\begin{figure}[t]
\vspace{-8mm}
	\centering
	\includegraphics[width = 0.85\linewidth]{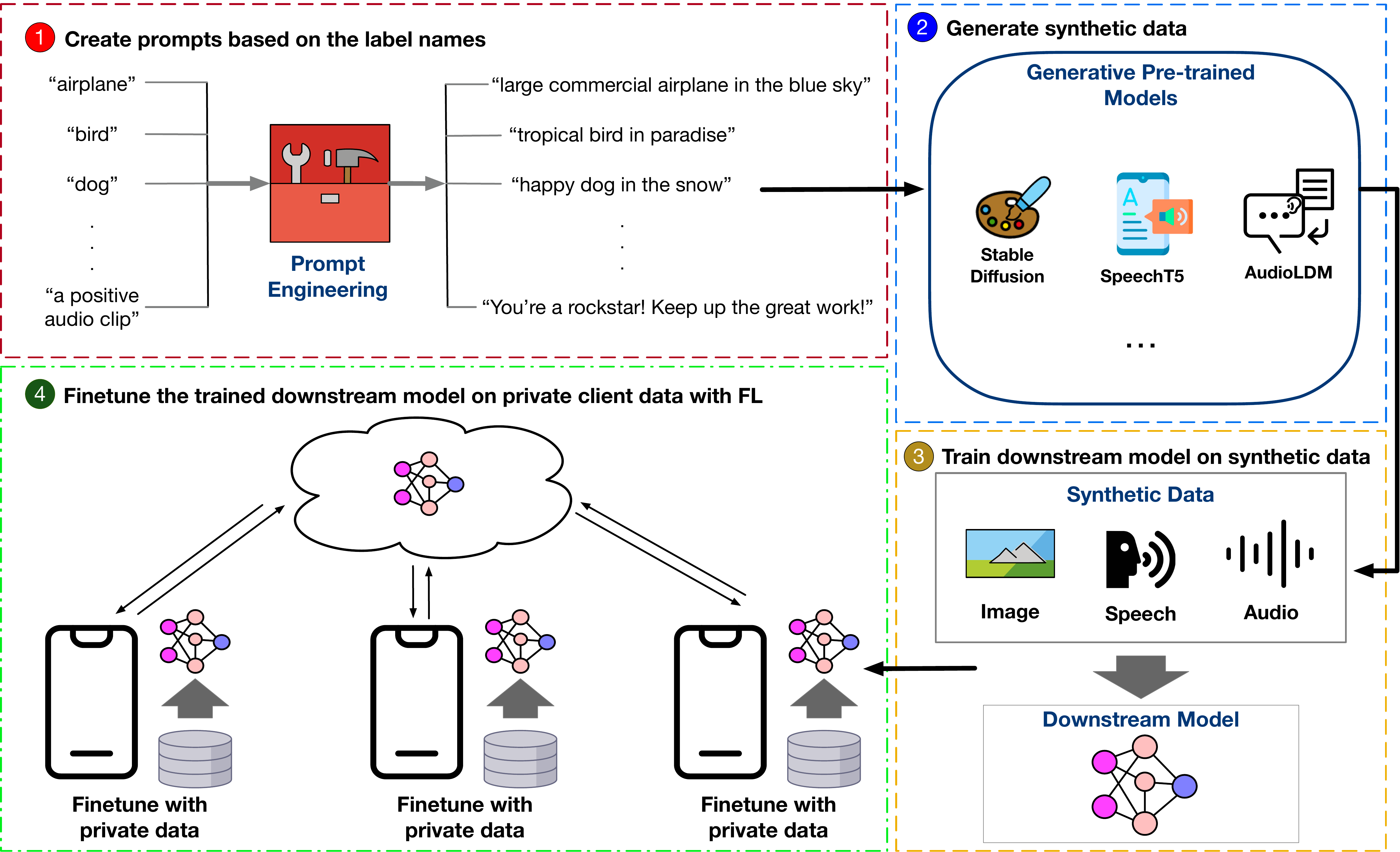}
    \caption{Overview of the proposed \texttt{GPT-FL} Framework.}
    \label{fig:fedgpt}
    \vspace{-5mm}
\end{figure}

The overall architecture of \texttt{GPT-FL} is illustrated in Figure~\ref{fig:fedgpt}. 
The objective of \texttt{GPT-FL} is to transfer the knowledge inside the pre-trained foundational models to the FL system to improve the performance of FL.
To achieve this objective, \texttt{GPT-FL} consists of four steps. 
First, prompts are created based on the label names at the server.
These prompts are then utilized to guide the generative pre-trained models to generate synthetic data. The server uses these generated synthetic data to train a downstream model and distributes the trained model to the clients. 
Lastly, the clients use the trained model as the starting point, and finetune the model with their private data under the standard FL framework.
The pseudocode of \texttt{GPT-FL} is presented in Appendix~\ref{algorithm}.
We describe the details in each step below. 

\subsection{Create Prompts based on Label Names}

As the first step of \texttt{GPT-FL}, a prompt that describes the desired content of the data is required to guide the synthetic data generation process. To do so, \texttt{GPT-FL} requires the clients to provide the set of label names\footnote{To protect user data privacy in FL setting, \texttt{GPT-FL} only requests the set of distinct label names instead of detailed label name distributions, and generates a uniform number of prompts for each label name.} of their private local data to generate prompts. 
However, prior research~\cite{Shipard2023DiversityID, He2022IsSD} shows that using only label names to generate prompts could restrict the quality and diversity of the generated synthetic data. Moreover, in FL, the server does not have access to detailed descriptions of the private data. To address these issues, \texttt{GPT-FL} incorporates LLMs such as GPT-3 to expand each input class's details and use them as prompts for synthetic data generation. 
As an example, for the label name "airplane", \texttt{GPT-FL} uses the following query for the LLM to generate the prompt as follows:

\begin{adjustwidth}{1cm}{1cm}
\vspace{-1.5mm}
\begin{lstlisting}[breakatwhitespace=true]
Q: " _ _ _ _ airplane _ _ _ _" Please fill in the blank and make it as a prompt to generate the image
A: Large commercial airplane in the blue sky.
\end{lstlisting}
\vspace{-2mm}
\end{adjustwidth}

Moreover, inspired by~\cite{Shipard2023DiversityID}, we randomly set the unconditional guidance scale of the Stable Diffusion model between 1 and 5 to further enrich the data diversity. In addition to the aforementioned techniques, it is worth noting that \texttt{GPT-FL} is flexible and compatible with other prompt engineering techniques that can be used to generate diversified synthetic data.

\textbf{Integration of Invertible Bloom Lookup Tables (IBLT) to Enhance Label Privacy.}
\texttt{GPT-FL} can employ IBLT to encode label names before sending them to the server so that the label information of each client is not leaked to the server~\cite{Gascn2023FederatedHH}. Specifically, each client locally encodes its unique label names into IBLT~\cite{Goodrich2011InvertibleBL}, a probabilistic data structure that can encode items in an open domain efficiently. The server linearly aggregates these IBLTs via the secure aggregation~\cite{SA} and decodes the aggregated table for the union of unique label names without revealing individual label information. More details about IBLT in \texttt{GPT-FL} are provided in Appendix~\ref{IBLT} via an illustrative experiment.

\vspace{-1mm}
\subsection{Generate Synthetic Data from Generative Pre-trained Model}
\vspace{-1mm}

Next, the generated prompts are used as the inputs to the generative pre-trained models to generate synthetic data.
In this work, we utilize the state-of-the-art Latent Diffusion Model~\cite{Rombach2021HighResolutionIS} loaded with Stable Diffusion V2.1 weights to generate synthetic images for image-based FL applications; and we utilize the state-of-the-art SpeechT5 model~\cite{Ao2021SpeechT5UE} for text-to-speech and AudioLDM model~\cite{Liu2023AudioLDMTG} for text-to-audio to generate synthetic speech and audio data, respectively.

It should be noted that the proposed \texttt{GPT-FL} is a general framework that supports other generative pre-trained models and data modalities beyond images and audio.
Notably, \texttt{GPT-FL} treats the generative pre-trained models as a service provider, utilizing only their APIs for synthetic data generation without modifying any internal parameters or structures. This design concept enables users to efficiently develop customized AI models without deploying the foundation model on the server side, making it compatible with the current trend of constructing downstream applications with API access to generative pre-trained models. 
This design feature enables \texttt{GPT-FL} to be widely applicable to a range of pre-trained models, facilitating flexible and efficient synthetic data generation for FL.

\subsection{Train Downstream Model on Generated Synthetic Data}

With the generated synthetic data, \texttt{GPT-FL} trains a downstream  model on the server in a centralized manner, and distributes the trained model to the clients participated in FL. This trained model acts as the initialized model for the following federated training process.
%
One note should be emphasized from our empirical experiences is that training with synthetic data is prone to overfitting, as synthetic data tend to be highly patternized compared to real data. To mitigate the effects of overfitting, we adopt relatively large weight decay hyperparameters and small learning rates compared to training with real data. The detailed hyper-parameter selections are listed in Appendix~\ref{hyper}.

\subsection{Finetune Trained Downstream Model on Private Client Data with FL}

Lastly, the clients use the trained model distributed from the server as the starting point, and finetune the model with their private data under the standard FL framework until the finetuning converges.
As such, \texttt{GPT-FL} does not alter the standard FL framework, making it fully compatible with secure aggregation protocols as in standard FL methods.
More importantly, 
%
unlike existing generated data-based approaches~\cite{Zhang2022FinetuningGM, Zhu2021DataFreeKD, pi2022dynafed}, \texttt{GPT-FL} does not introduce any additional hyper-parameters beyond the standard FL framework. This significantly simplifies the hyper-parameter optimization process, making \texttt{GPT-FL} much more practically useful.


%% file: theory.tex
\section{Connection to Theory} \label{sec:theory}
In this section, we theoretically prove that pre-training on synthetic data could accelerate FL training and improve generalization performance, which are the main reasons for the effectiveness of \texttt{GPT-FL}.
In empirical risk minimization problems, there is a general assumption that the training dataset is a randomly drawn subset of the global data. Thus, the artificial data generated from pre-trained generative models can be considered as another random subset of the global data. From this perspective, we interpret the pre-training on the synthetic data as a model training biased toward the distribution of the synthetic data. 

We first define the optimal gradient as $\nabla F(x)$ and the stochastic gradient as $\nabla f(x)$. Under the conventional assumption, our stochastic gradient approximation is unbiased such that $\mathbb{E}[\nabla f(x)] = \nabla F(x)$. Similarly, we define the optimal gradient of the artificial data as $\nabla F'(x)$ and the corresponding stochastic gradient as $\nabla f'(x)$. Likewise, the stochastic approximation is unbiased, so $\mathbb{E}[\nabla f'(x)] = \nabla F'(x)$. Under these settings, even though the datasets are assumed to be randomly drawn from the same global data, $\mathbb{E}[\nabla f'(x)] = \nabla F'(x) \neq \nabla F(x) = \mathbb{E}[\nabla f(x)]$. We quantify this difference based on the conventional biased gradient setting~\cite{Driggs2019OnBS, Killen2021NonlinearP} as follows:

\vspace{-2mm}
\begin{equation}
\nabla f'(x) = \nabla f(x) + b(x),
\end{equation}
\vspace{-3mm}
\begin{equation}
\|b(x)\|^2 \leq m\|\nabla f(x)\|^2 + \zeta^2, \quad \forall x \in \mathbb{R}^d,
\end{equation}

where $b(x)$ is a bias, $b: \mathbb{R}^d \to \mathbb{R}^d$, and $m$ and $\zeta$ are small constants. It has already been shown that the biased gradient still guarantees convergence for non-convex and smooth problems~\cite{Driggs2019OnBS, Killen2021NonlinearP, Ajalloeian2020OnTC}. Specifically, biased gradients lead the model to a neighboring region of the local minimum. Thus, the pre-training will most likely make the main training begin with a low initial loss.
Notably, since $\mathbb{E}[\nabla F'(x)] = \mathbb{E}[\nabla F(x)] = \nabla \overline{F}(x)$, where $\nabla \overline{F}(x)$ is the optimal gradient computed from the global data, the synthetic dataset is expected to have lower variance as its size increases based on the central limit theorem. 
Given a fixed training dataset, a larger synthetic dataset leads to a lower degree of bias (smaller $m$ and $\zeta^2$). It has been shown that a lower degree of bias results in faster convergence~\cite{Killen2021NonlinearP}.

In addition to training acceleration by achieving a lower initial loss, pre-training on synthetic data is expected to improve generalization performance. Generalization performance is defined as the loss gap between $\overline{F}(x) - F(x)$ in previous works~\cite{Barnes2022ImprovedIG, Sun2023UnderstandingGO, Gholami2024ImprovedGB}. Informally, as our model is trained not only on the training dataset but also on the synthetic data, the loss gap will likely be reduced compared to models trained solely on the training dataset. We will empirically demonstrate this and analyze the benefits of \texttt{GPT-FL} in the experimental section.

%% file: 4_exp.tex
\section{Experiments} \label{sec:experiment}

\textbf{Datasets, Models, and Tasks.} We assess \texttt{GPT-FL}'s performance across five datasets from three FL applications: image classification, speech keyword spotting, and environmental sound classification.
In image classification, we experiment with CIFAR-10, CIFAR-100~\cite{Krizhevsky2009LearningML}, and Oxford 102 Flower~\cite{Nilsback2008AutomatedFC} datasets. We utilize ConvNet~\cite{pi2022dynafed}, ResNet18, ResNet50~\cite{He2015DeepRL}, and VGG19~\cite{Simonyan2014VeryDC} models. CIFAR-10 and CIFAR-100 feature diverse objects, while Oxford 102 Flower provides high-resolution images for fine-grained analysis.
%
For audio tasks, we utilize the Google Command dataset~\cite{Warden2018SpeechCA} for keyword spotting and the ESC-50 dataset~\cite{piczak2015dataset} for environmental sound classification, employing models from prior research~\cite{Zhang2022FedAudioAF}.
%
More details about the data and model setups is described in the Appendix~\ref{data}.

\textbf{Data Heterogeneity.} CIFAR-10 and CIFAR-100 are heterogeneously partitioned among 100 clients using the Dirichlet distribution $Dir_K(\alpha)$ with $\alpha=0.1$ and $\alpha=0.5$, respectively, following prior studies~\cite{Cho2022HeterogeneousEK}. For the smaller Flowers102 dataset, we similarly partition it into 50 subsets. Google Speech Command dataset is inherently non-IID, distributed across 2,618 speaker IDs. Following the previous work~\cite{Zhang2022FedAudioAF}, we partition the ESC-50 dataset into 100 subsets using $Dir_K(\alpha)$ where $\alpha=0.1$.
\textbf{Baselines and Evaluation Metrics.} We compare \texttt{GPT-FL} against three categories of baselines: 1) standard FL methods without the use of public or generated synthetic data -- FedAvg, FedProx, and Scaffold; 2) FL methods that involve the use of public data -- FedDF, DS-FL, and Fed-ET; and 3) FL methods that utilize generated synthetic data -- FedGen and DynaFed\footnote{We did not compare with FedFTG because its code is not open-source, and we could not reproduce their results following the paper.}. 
We use the test accuracy of the trained model as our evaluation metric.
We run experiments with three different random seeds and report the average and standard deviation. The details of the hyper-parameter selection of each dataset and experiment are described in Appendix~\ref{hyper}.

\begin{table}[t]
\vspace{-10mm}
\centering
\caption{Model accuracy comparison between \texttt{GPT-FL} and existing FL methods. For public data-based methods FedDF, DS-FL and Fed-ET, the results on CIFAR-10 and CIFAR-100 are obtained from~\cite{Cho2022HeterogeneousEK}, and the results on Flowers102 are marked as N/A given the practical challenge on finding a set of  suitable public data that can boost its performance.}
\label{tab:overall_performance}
\vspace{2mm}
\resizebox{1.0\textwidth}{!}{%
\begin{tabular}{@{}ccccccccccc@{}}
\toprule
& \multirow{2}{*}{\textbf{Method}} & \multirow{2}{*}{\begin{tabular}[c]{@{}c@{}}\textbf{Training} \\ \textbf{Model} \end{tabular}} & & \multicolumn{3}{c}{\textbf{High Data Heterogeneity} ($\mathbf{\alpha = 0.1}$)} &  & \multicolumn{3}{c}{\textbf{Low Data Heterogeneity} ($\mathbf{\alpha = 0.5}$)}\\
\cmidrule(lr){4-11} &  & &  & \textbf{CIFAR-10} & \textbf{CIFAR-100} & \textbf{Flowers102} &  & \textbf{CIFAR-10} & \textbf{CIFAR-100} & \textbf{Flowers102}\\ \midrule
& FedAvg & \multirow{3}{*}{VGG19} &  & 71.19 (\textpm\;0.27) & 30.21   (\textpm\;0.32) & 30.30 (\textpm\;0.16) & & 74.82 (\textpm\;0.23) & 33.12   (\textpm\;0.13) & 34.75 (\textpm\;0.90) \\
 & FedProx &  &  & 72.45 (\textpm\;0.13) & 31.51 (\textpm\;0.11) & 33.23 (\textpm\;0.24) &  & 75.24 (\textpm\;0.19) & 33.64 (\textpm\;0.08) & 40.56 (\textpm\;0.19) \\
& SCAFFOLD &  &  & 75.12 (\textpm\;0.20) & 30.61 (\textpm\;0.57) & 26.75 (\textpm\;0.50) &  & 78.69 (\textpm\;0.15) & 34.91 (\textpm\;0.61) & 33.21 (\textpm\;0.41) \\
\midrule
& FedDF & \multirow{3}{*}{VGG19} & & 73.81 (\textpm\;0.42) & 31.87   (\textpm\;0.46) & N/A &  & 76.55 (\textpm\;0.32) & 37.87   (\textpm\;0.31) & N/A \\
& DS-FL & & & 65.27 (\textpm\;0.53) & 29.12 (\textpm\;0.51) & N/A &  & 68.44 (\textpm\;0.47) & 33.56 (\textpm\;0.55) & N/A\\
& Fed-ET & & & 78.66 (\textpm\;0.31) & 35.78 (\textpm\;0.45) & N/A &  & 81.13 (\textpm\;0.28) & 41.58 (\textpm\;0.36) & N/A\\ 
\midrule
& FedGen & \multirow{2}{*}{ConvNet \tablefootnote{\cite{Zhu2021DataFreeKD,pi2022dynafed} only reported results on ConvNet. We tested these two methods on VGG19 but they are not converged.}} &  & 42.05 (\textpm\;0.93) & 26.64 (\textpm\;0.66) & Not Converged &  & 54.86 (\textpm\;0.13) & 34.03 (\textpm\;0.42) & Not Converged \\
& DynaFed &  &  & 71.59 (\textpm\;0.10) & 36.08 (\textpm\;0.15) & Not Converged &  & 75.66 (\textpm\;0.21) & 43.82 (\textpm\;0.30) & Not Converged \\
\midrule
& \multirow{2}{*}{\textbf{GPT-FL}} & VGG19 &  & \textbf{82.16 (\textpm\;0.13)} & \textbf{47.80 (\textpm\;0.32)} & \textbf{70.56 (\textpm\;0.34)} &  & \textbf{82.17 (\textpm\;0.20)} & \textbf{48.39 (\textpm\;0.17)} & \textbf{74.84 (\textpm\;0.43)} \\
& & ConvNet & & \textbf{72.62 (\textpm\;0.24)} & \textbf{42.66 (\textpm\;0.19)} & \textbf{37.91 (\textpm\;0.43)} & &  \textbf{77.18 (\textpm\;0.21)} & \textbf{47.89 (\textpm\;0.28)} & \textbf{48.61 (\textpm\;0.51)} \\
\bottomrule
\end{tabular}%
}
\vspace{-5mm}
\end{table}

\subsection{Performance Comparison with State-of-the-Art FL Methods}

First, we compare the performance of \texttt{GPT-FL} with state-of-the-art FL methods. 
To enforce fair comparisons, in this experiment, we choose to evaluate on the three image datasets since baseline methods FedGen and DynaFed only support image data. 
Moreover, we used the same models (VGG19 and ConvNet) and experiment settings as previous work~\cite{Cho2022HeterogeneousEK, pi2022dynafed}. 
In each round, we randomly sample 10 clients from 100 clients for CIFAR and use all 50 clients for Flowers102. We choose FedAvg as the FL optimizer. All the training starts from random initialization and total number of communication rounds is set to 500. The local training epoch is 1 for all experiments.

\textbf{Overall Performance.} 
Table~\ref{tab:overall_performance} summarizes our results.
We make three key observations:
(1) \texttt{GPT-FL} consistently outperforms all the baselines we selected in Table~\ref{tab:overall_performance} across all three datasets.
(2) In direct comparison with state-of-the-art generated data-based FL methods, although FedGen and DynaFed perform reasonably well on CIFAR-10/100, they do not converge on Flowers102 whose images have higher resolutions than CIFAR. Moreover, both FedGen and DynaFed fail to converge when training a larger VGG19 model on Flowers102 and even lower-resolution CIFAR-10/100. In contrast, \texttt{GPT-FL} not only converges but also achieves state-of-the-art accuracy on Flowers102. More importantly, \texttt{GPT-FL} is able to support larger model, and its accuracy is significantly higher than the smaller ConvNet.
%
%
(3) For Flowers102, as both public data-based and generated data-based FL methods are confronted with challenges, the only viable options are standard FL methods and \texttt{GPT-FL}. As shown, with the same model, \texttt{GPT-FL} outperforms standard FL methods by a significant margin. 
%
%

\begin{figure}
  \vspace{-10mm}
  \begin{minipage}[b]{.32\textwidth}
  \centering
  \includegraphics[width=\textwidth]{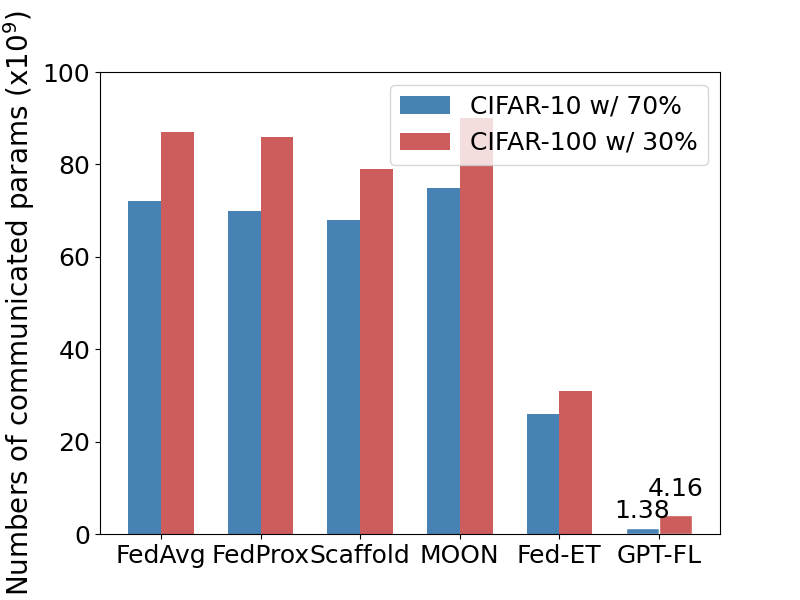}
  \caption{Communication costs of standard FL methods, public data-based methods and \texttt{GPT-FL} to achieve the target test accuracy.}
  \label{fig:comm}
  \end{minipage}%
  \hfill
  \begin{minipage}[b]{.32\textwidth}
  \centering
  \includegraphics[width=\textwidth]{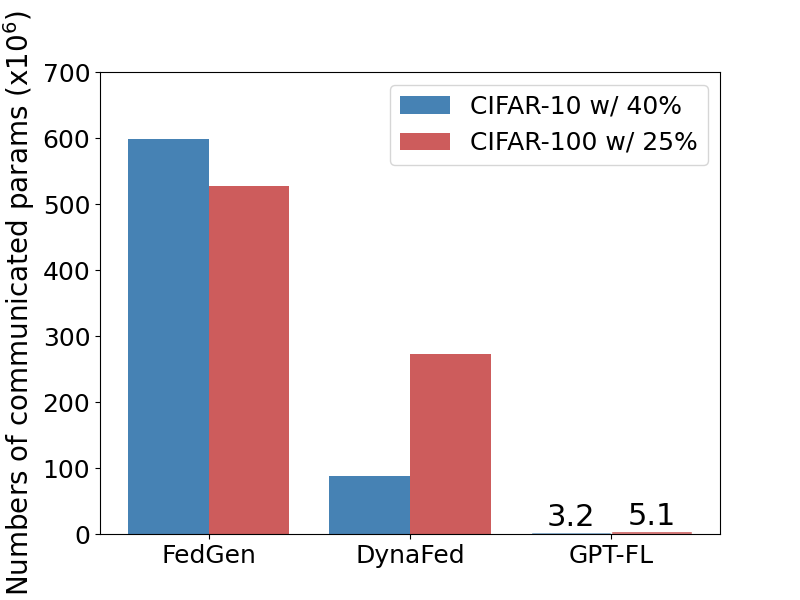}
  \caption{Communication costs of generated data-based methods and \texttt{GPT-FL} to achieve the target test accuracy.}
  \label{fig:comm_syn}
  \end{minipage}%
  \hfill
  \begin{minipage}[b]{.32\textwidth}
  \centering
  \includegraphics[width=\textwidth]{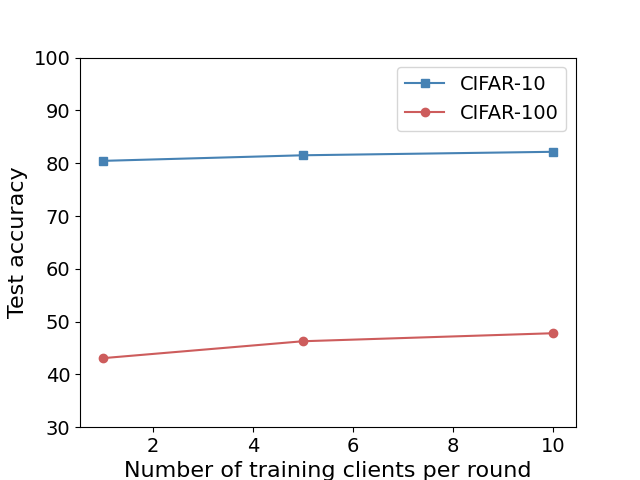}
  \caption{Test accuracy of \texttt{GPT-FL} for CIFAR-10/100 under different client sampling rates.}
  \label{fig:comp_sample}
  \end{minipage}%
\end{figure}


\textbf{Communication Efficiency.}
In addition to model accuracy, we evaluate the communication costs of \texttt{GPT-FL} compared to exisiting FL methods on CIFAR-10/100 under high data heterogeneity. The communication cost is quantified by the total number of model parameters exchanged between the server and the clients to achieve a target test accuracy. Figures~\ref{fig:comm} and \ref{fig:comm_syn} illustrate the comparisons with standard/public data-based FL\footnote{We do not compare with FedDF and DS-FL as they do not achieve competitive model accuracy.} and generative data-based methods under VGG19 and ConvNet, respectively. The target test accuracies in Figure~\ref{fig:comm_syn} are set to be lower given the low accuracies achieved by FedGen. Notably, \texttt{GPT-FL} reduces communication costs by up to 94\% against the best public data-based method, Fed-ET, and 98\% against the best generative data-based method, DynaFed. These results highlight the significant advantage of \texttt{GPT-FL} in communication reduction over state-of-the-art FL methods.

Furthermore, we examine the client sampling rate's impact, a crucial hyper-parameter in FL, on model performance. As depicted in Figure~\ref{fig:comp_sample}, \texttt{GPT-FL} achieves 80.44\% and 43.07\% test accuracies on CIFAR-10 and CIFAR-100, respectively, with only one client per round. This surpasses other FL methods that use nine times more clients per round. These findings underscore \texttt{GPT-FL}'s superior client sampling efficiency, making it highly effective in scenarios with limited client participation.


\subsection{Understanding GPT-FL} \label{Sec:ablation}

\textbf{(1) Can we only rely on centralized training with synthetic data?}

\begin{table*}[t]
\vspace{-2mm}
\caption{Performance of the generated downstream model and standard FL on benchmark datasets. "1x Synthetic" represents the size of synthetic data is one time as the real data.}
\label{syn_vs_fl}
\begin{center}
\scriptsize
\resizebox{1.0\textwidth}{!}{
\begin{tabular}{ccccccc}
\toprule  
& \textbf{Dataset} & \textbf{1x Synthetic} & \textbf{2x Synthetic} & \textbf{3x Synthetic} & \textbf{FedAvg} & \textbf{FedOpt} \\
\cmidrule{1-7}
\multirow{3}{*}{Image Data} & CIFAR-10 & 61.48 (\textpm\;0.08) & 67.41 (\textpm\;0.40) & \textbf{75.65 (\textpm\;0.09)} & 64.48 (\textpm\;0.13) & 72.68 (\textpm\;0.22)\\
\cmidrule{2-7}
& CIFAR-100 & 24.70 (\textpm\;0.00) & 33.41 (\textpm\;0.01) & \textbf{41.76 (\textpm\;0.03)} & 25.89 (\textpm\;0.67) & 20.85 (\textpm\;0.14) \\
\cmidrule{2-7}
& Flowers102 & 24.94 (\textpm\;0.57) & 28.26 (\textpm\;0.14) & \textbf{31.29 (\textpm\;0.18)} & 30.30 (\textpm\;0.16) & 26.43 (\textpm\;0.09)\\
\cmidrule{1-7}
\multirow{2}{*}{Audio Data} & Google Command & 24.78 (\textpm\;0.04) & 25.65 (\textpm\;0.07) & 26.24 (\textpm\;0.01) & 73.68 (\textpm\;0.49) & \textbf{83.01 (\textpm\;0.23)}\\
\cmidrule{2-7}
& ESC-50 & 6.89 (\textpm\;0.29) & 8.68 (\textpm\;0.35) & 12.72 (\textpm\;0.31) & 22.76 (\textpm\;1.01) & \textbf{32.49 (\textpm\;0.57)}\\
\cmidrule{1-7}
Text Data & MELD & 58.94 (\textpm\;0.55) & 59.36 (\textpm\;0.51) & 59.38 (\textpm\;0.36) & 57.91 (\textpm\;0.15) & \textbf{61.88 (\textpm\;0.22)}\\
\bottomrule
\end{tabular}
}
\end{center}
\label{tab:syn_vs_fl}
\vspace{-7mm}
\end{table*}

To answer this question, we compare the model performance between generated downstream model by centralized training with synthetic data and the global model by standard FL training with private data on both image and audio benchmark datasets. Different from the previous section, we select ResNet18 for CIFAR-10 and Flowers102, and ResNet50 for CIFAR-100. 
We also perform text-related experiments using the MELD dataset~\cite{poria2019meld} for sentiment analysis. Further details about these experiments are provided in the Appendix~\ref{sec:meld}.
We report the best F1 score for the audio and text datasets. The results are summarized in Table~\ref{tab:syn_vs_fl}.

\textbf{Impact of Out-of-Domain Data Generation.} 
We choose the ESC-50 and Google Speech Commands datasets to examine the impact of out-of-domain data generation for the generative pre-trained model. We did not conduct a similar analysis for the image or text datasets as the LAION-5B~\cite{Schuhmann2022LAION5BAO} open-source dataset for training the Stable Diffusion model we used is a vast collection of publicly available datasets, including nearly all relevant ones for our experiments. We also include experiments regarding the medical images and food images in Appendix~\ref{app:covid} for a comprehensive study.

Our results show that synthetic image outperforms synthetic audio in model performance under centralized training. We observed that centralized training with synthetic images achieves higher accuracy than FL setups for all three image benchmark datasets. In contrast, centralized training with synthetic audio performs worse than FL setups for both datasets. The finding from the audio experiments aligns with the previous study \cite{li2018training} that utilizes pure synthetic speech data to train the automatic speech recognition system leading to substantial performance degradation. One plausible explanation is related to the relatively small training data sizes (approximately 400M sentences) and constrained domain knowledge (book corpus) compared to training other generative pre-trained models like Stable Diffusion. Further analysis of synthetic audio quality is detailed in Appendix~\ref{audio data}.

\begin{figure}[t!]
\vspace{-10mm}
  \begin{minipage}[b]{.32\textwidth}
  \centering
  \includegraphics[width=\textwidth]{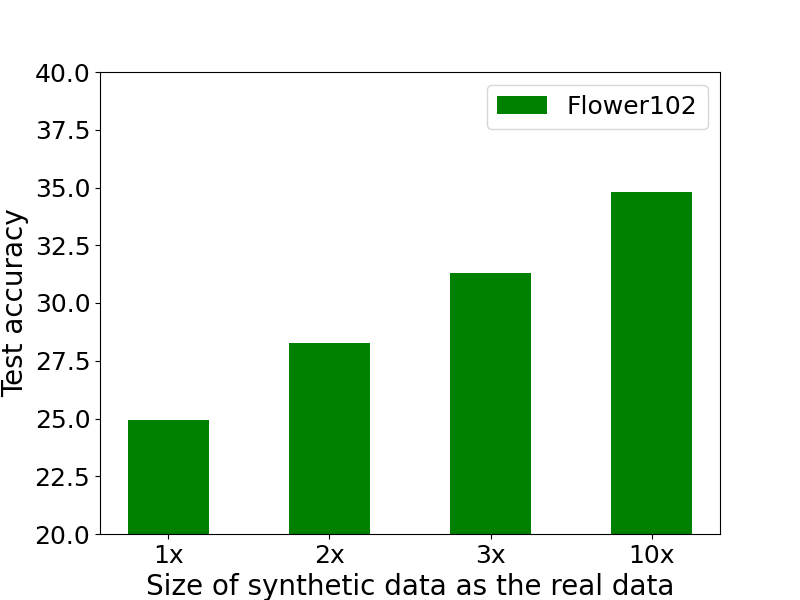}
  \caption{Impact of synthetic data sample number to the generated downstream model.}
  \label{fig:flower}
  \end{minipage}%
  \hfill
  \begin{minipage}[b]{.32\textwidth}
  \centering
  \includegraphics[width=\textwidth]{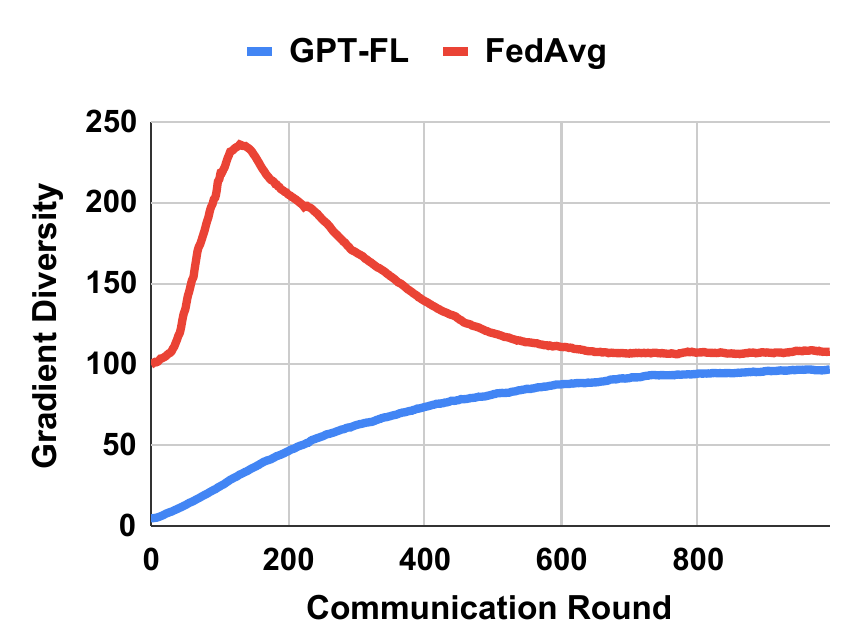}
  \caption{Smoothed Gradient diversity of client updates during training on Google speech commands dataset.}
  \label{fig:gradient}
  \end{minipage}%
  \hfill
  \begin{minipage}[b]{.32\textwidth}
  \centering
  \includegraphics[width=\textwidth]{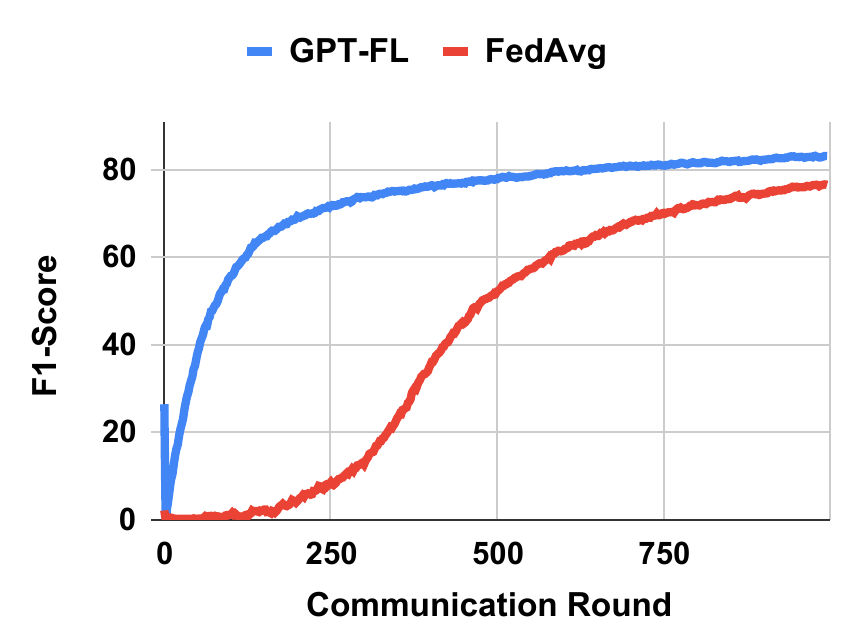}
  \caption{Learning curve of the global model during training on Google speech commands dataset.}
  \label{fig:loss}
  \end{minipage}%
  \vspace{-5mm}
\end{figure}

\textbf{Impact of Synthetic Data Volume.} 
%
The results show that centralized training benefits from an increase in synthetic data volume. We explored this effect on the Flowers102 dataset by augmenting the synthetic data to tenfold the size of the real data. As illustrated in Figure~\ref{fig:flower}, model performance improves with the expansion of synthetic data.
One justification for this finding is that enlarging the number of synthetic data enriches the diversity and increases overlap between the synthetic and real data, allowing the model to learn more robust and generalizable features. 
The results also validate our theoretical analysis in Section~\ref{sec:theory}.

\textbf{(2) Why \texttt{GPT-FL} needs to fine-tune the downstream models federatively?}

\textbf{Evaluating Client-isolated Fine-tuning against \texttt{GPT-FL} Performance.}
To underscore the effectiveness of federation in fine-tuning, we compare the performance of local fine-tuning in isolation against FL fine-tuning. 
We select the VGG19 model for the CIFAR-100 dataset to align with Table~\ref{tab:overall_performance}. 
In the isolated fine-tuning scenario, we select 10 clients at random, allowing each to fine-tune the synthetic-data-based downstream model independently with its local data for 500 epochs. The average accuracy from these clients is then computed. For the \texttt{GPT-FL} setup, we maintain the experimental parameters as per the setups described in Table~\ref{tab:fedgpt}. Results are summarized in Table~\ref{tab:local}. 
%
Additionally, an ablation study detailed in Appendix~\ref{distribution} investigates the impact of distribution drift in synthetic data on \texttt{GPT-FL} performance, highlighting its effectiveness under varied data distributions.

\begin{table*}[h]
\vspace{-3mm}
\caption{Accuracy performance comparison between locally fine-tune in isolation and \texttt{GPT-FL}.}
\vspace{-2mm}
\begin{center}
\scriptsize
\resizebox{0.7\textwidth}{!}{
\begin{tabular}{ccc}
\toprule  
\textbf{Dataset} & \textbf{Locally Fine-tune in Isolation} & \textbf{\texttt{GPT-FL} w/ FedAvg}\\
\cmidrule{1-3}
CIFAR-100 & 35.53\% (\textpm\;0.57) & \textbf{47.80\% (\textpm\;0.32)}\\
\cmidrule{1-3}
Google Speech Command & 23.00\% (\textpm\;0.13) & \textbf{81.90\% (\textpm\;0.20)} \\
\bottomrule
\end{tabular}
}
\end{center}
\label{tab:local}
\end{table*}

The results show that fine-tuning in isolation at the client level yields significantly lower accuracy compared to the \texttt{GPT-FL} approach, which fine-tunes synthetic-data-based downstream models federatively using private data. The primary reason for this disparity is the limited amount and skewed label distribution of the local data available to each client, which is insufficient for individually tuning the model to achieve optimal performance. 
These findings clearly demonstrate the value of FL in fine-tuning, especially given the limitations of local data in terms of volume and diversity.

\textbf{(3) What benefits does \texttt{GPT-FL} bring?}

In this section, we want to examine how fine-tuning the downstream models federatively can lead to performance improvements. 
We evaluate its performance with both FedAvg and FedOpt as the server aggregator. 
Our experimental results are presented in Table~\ref{tab:fedgpt}. We also conduct experiments demonstrating \texttt{GPT-FL}'s compatibility with existing FL strategies and pre-trained models, confirming that it enhances existing methods without altering their structures. Detailed analyses and results are provided in the Appendix~\ref{harmonization} due to space constraints.

\begin{table*}[h]
\caption{Accuracy comparison between generated downstream model, standard FL and \texttt{GPT-FL}. 
"$\Delta$Metric" represents the accuracy increment by \texttt{GPT-FL} on top of the generated downstream model.}
\begin{center}
\scriptsize
\resizebox{1.0\textwidth}{!}{
\begin{tabular}{ccccccc}
\toprule  
\textbf{Dataset} & \textbf{3x Synthetic} & \textbf{FedAvg} & \textbf{FedOpt} & \textbf{\texttt{GPT-FL} w/ FedAvg} & \textbf{\texttt{GPT-FL} w/ FedOpt} & \textbf{$\Delta$Metric}\\
\cmidrule{1-7}
CIFAR-10 & 75.65 (\textpm\;0.09) & 64.48 (\textpm\;0.13) & 72.68 (\textpm\;0.22) & \textbf{81.38 (\textpm\;0.05)} & 79.08 (\textpm\;0.17) & $\mathbf{\uparrow}$ \textbf{5.73}\\
\cmidrule{1-7}
CIFAR-100 & 41.76 (\textpm\;0.03) & 25.89 (\textpm\;0.67) & 20.85 (\textpm\;0.14) & \textbf{62.83 (\textpm\;0.31)} & 48.80 (\textpm\;0.12) & $\mathbf{\uparrow}$ \textbf{21.07}\\
\cmidrule{1-7}
Flowers102 & 31.29 (\textpm\;0.18) & 30.30 (\textpm\;0.16) & 26.43 (\textpm\;0.09) & 70.56 (\textpm\;0.34) & \textbf{77.57 (\textpm\;0.03)} & $\mathbf{\uparrow}$ \textbf{46.28}\\
\cmidrule{1-7}
Google Command & 26.24 (\textpm\;0.01) & 73.68 (\textpm\;0.49) & 83.01 (\textpm\;0.23) & 81.90 (\textpm\;0.20) & \textbf{83.46 (\textpm\;0.11)} & $\mathbf{\uparrow}$ \textbf{57.22}\\
\cmidrule{1-7}
ESC-50 & 12.72 (\textpm\;0.31) & 22.76 (\textpm\;1.01) & 32.49 (\textpm\;0.57) & 41.80 (\textpm\;0.32) & \textbf{43.46 (\textpm\;0.30)} & $\mathbf{\uparrow}$ \textbf{30.74}\\
\cmidrule{1-7}
MELD & 59.38 (\textpm\;0.36) & 57.91 (\textpm\;0.15) & 61.88 (\textpm\;0.22) & 64.48 (\textpm\;0.16) & \textbf{65.70
 (\textpm\;0.29)} & $\mathbf{\uparrow}$ \textbf{6.32}\\
\bottomrule
\end{tabular}
}
\end{center}
\label{tab:fedgpt}
\end{table*}

\textbf{Effectiveness of Private Data.}
Our experiments demonstrate the effectiveness of incorporating private data with FL into the finetuning process of the downstream model generated from synthetic data. As shown in Table~\ref{tab:fedgpt}, regardless of the modality and quality of the synthetic data used to generate the downstream model, FL fine-tuning leads to significant performance gains, outperforming the ones trained solely with FL or CL combined with synthetic training by a large margin. Furthermore, we observe that fine-tuning with private data can especially benefit the cases for out-of-domain synthetic data, such as in the audio data. For example, \texttt{GPT-FL} with FedOpt could achieve 43.46 test accuracy in the ECS-50 dataset, which nearly provides two times increment than standard FL and three times increment than centralized training by synthetic data. These results suggest that leveraging private data with FL in the fine-tuning process can greatly enhance the performance of synthetic data-generated models, making them more suitable for real-world applications.

\textbf{Generated Downstream Model Helps FL Optimization.}
To gain a comprehensive understanding of why the custom models built using \texttt{GPT-FL} provide benefits to performance improvements, we decided to compare the gradient diversity between model weights initialized by \texttt{GPT-FL} and random initialization. Specifically, we apply the definition of the gradient diversity introduced from \cite{yin2018gradient} by adapting the gradients $g_{i}$ to client update $\Delta_{i}$:

\begin{equation}
    \Delta_{\textit{S}} = \frac{\sum_{i\in\textit{S}}||\Delta_{i}||^2}{||\sum_{i\in\textit{S}}\Delta_{i}||^2}
\end{equation}

where $S$ is the set of sampled clients in each communication round and $i$ represents the client index. 
As shown in Figure \ref{fig:gradient}, the gradient diversity plot for FedAvg reveals that \texttt{GPT-FL} displays lower initial gradient diversity compared to random initialization. Over training time, both \texttt{GPT-FL} and random initialization converge to similar gradient diversity levels, consistent with the performance curve in Figure \ref{fig:loss}, where a larger $\Delta_{S}$ corresponds to slower convergence rate. This aligns with prior findings \cite{Nguyen2022WhereTB, Chen2022OnTI} and our theoretical analysis in Section~\ref{sec:theory}, indicating that starting from a pre-trained model leads to less variation in local client updates, potentially addressing the client drift issue.



%% file: 5_conclusion.tex
\vspace{-2mm}
\section{Conclusion} \label{sec:conclusion}
\vspace{-2mm}
We present \texttt{GPT-FL}, a generative pre-trained model-assisted federated learning framework. \texttt{GPT-FL} leverages the generative pre-trained model to generate diversified synthetic data for a wide range of data modalities before FL training. This synthetic data is then utilized to construct a downstream model, which undergoes fine-tuning with private data within a standard FL framework. Our experimental results showcase the remarkable performance of \texttt{GPT-FL} when compared to state-of-the-art FL methods. Moreover, through detailed ablation studies, we demonstrate that \texttt{GPT-FL} is a flexible and applicable framework solution to the challenges associated with cross-device FL scenarios.

\textbf{Limitations and Future works.} 
%
While we recognize the importance of theoretical analysis, the inherent complexity and opacity of current LLMs pose significant challenges to formulating comprehensive theoretical frameworks. Our proposed questions are pivotal in the context of leveraging LLMs and foundational models, which often present as "black-box" systems with limited theoretical support. Our detailed experimental analysis, therefore, serves as a foundational step toward understanding the interplay between synthetic data generation, model fine-tuning, and FL in the context of LLMs and foundation models. We believe these insights are invaluable for guiding future theoretical and empirical studies in this rapidly evolving field.

%% file: appendix.tex
\newpage
\section{Appendix} \label{Appendix}

\subsection{Algorithm Overview} \label{algorithm}
Consider the standard FL setting~\cite{McMahan2016CommunicationEfficientLO}, in which the FL system is composed of a server with $\mathcal{K}$ clients, whose data is only locally kept without sharing. The clients cooperate in training a global model $\mathcal{W}_g$ with parameters $\theta$ with the aim to solve the optimization problem formulated as follows
\begin{equation}
    \min\limits_{\theta} f(\theta) := \sum_{i=1}^{\mathcal{K}} p_i F_i(\theta)
\end{equation}
where $F_i(\cdot)$ represents the local objective of client $i$ and $p_i$ denotes the aggregation weight of client $i$ satisfying $p_i \geq 0$ and $\sum_{i=1}^{\mathcal{C}} p_i=1 $.

\begin{algorithm}[h!]
\SetKwInput{Input}{Input}
\SetKwInput{Output}{Output}
    \Input{$\mathcal{G}$: generative pre-trained model, $T$: FL communication rounds}
    // Generate downstream model weight $\theta_{init}$ from $\mathcal{G}$ on server $S$ \\
    \For{each client $i \in \{1, \cdots, K \}$ \textbf{in parallel}}{
        submit label names set $y_i$ to server $S$ \\
    }
    global label name set $y_s \leftarrow$ aggregate $\langle{y_i}\rangle$ \\
    synthetic dataset $D^\prime \leftarrow \mathcal{G}(y_s)$\\
    $\theta_{init} \leftarrow$ centralized training $D^\prime$ on server $S$\\
    // Finetune trained downstream model on private client data with FL \\
    Server $S$ sets ${\theta}^{0}_{s}$ =  $\theta_{init}$ \\
    \For{$r \in \{0, \cdots, T-1 \}$ \textbf{communication rounds}}{
        Sample $n$ clients uniformly at random to define $\mathcal{C}$, and send ${\theta}^{r}_{s}$ to clients in $\mathcal{C}$ \\
        {\textbf{Clients $c \in\mathcal{C}$ in parallel do:}}\\
            \hspace*{1em} ${\theta}^{r}_{c}$ $\leftarrow$ local model update \\
        {\textbf{Server S do:}}\\
            \hspace*{1em} ${\theta}^{r+1}_{s}$ $\leftarrow$ aggregate $\langle{{\theta}^{r}_{c}}\rangle$
    }
    \Output{${\theta}^{T}_{s}$}
\caption{
    GPT-FL.
}
\label{alg:gptfl}
\end{algorithm}

\subsection{Integration of Invertible Bloom Lookup Tables (IBLT) to Enhance Label Privacy} \label{IBLT}

It should be noted that \texttt{GPT-FL} can employ Invertible Bloom Lookup Tables (IBLT) to encode label names before sending them to the server so that the label information of each client is not leaked to the server~\cite{Gascn2023FederatedHH}. Specifically, each client locally encodes its unique label names into IBLT, a probabilistic data structure that can encode items in an open domain efficiently. The server linearly aggregates these IBLTs via the secure aggregation~\cite{SA} and decodes the aggregated table for the union of unique label names without revealing individual label information. 

Within the \texttt{GPT-FL} framework, the set of distinct label names is sourced from an open domain. The server lacks detailed length information on the set, making it challenging to directly encode the label names properly for secure aggregation. To address this, we propose to locally encode the unique label names into IBLT~\cite{Goodrich2011InvertibleBL} data structure, a randomized data structure efficient in storing key-value pairs within an open domain. IBLT is a bloom filter-type linear data structure that supports the efficient listing of inserted elements and their precise counts, with table size scaling linearly with unique keys. IBLT sketches are amenable to linear summation, thus compatible with secure aggregation protocols.

In the \texttt{GPT-FL} framework's IBLT integration, each client locally encodes its distinct label names into IBLT and transmits it to the server. The server performs linear aggregation of these IBLTs through a secure multi-party computation protocol, subsequently decoding the aggregated table to obtain total label name counts without revealing individual label information. By leveraging the collective label name histogram, the server determines the union of distinct label names for data generation, maintaining the privacy of client-specific details. This approach finds validation in prior research~\cite{Gascn2023FederatedHH}, where IBLT demonstrated its efficacy in addressing private heavy hitters within federated analytics.

To better demonstrate the integration of IBLT in \texttt{GPT-FL}, we provide an illustrated experiment as an example. The experiment is conducted with the TensorFlow Federated IBLT API~\cite{TFFIBLT}. We partition the CIFAR-10 dataset heterogeneously amongst 100 clients using the Dirichlet distribution $Dir_K(\alpha)$ with $\alpha$ equal to 0.1. As the server does not know the length of the dataset initially, we set the capacity of the IBLT sketch to 50, which is much larger than the total number of unique labels inside CIFAR-10 (i.e., 10). Each client encodes its unique set of label names into IBLT and sends it to the server. The server would aggregate them via the secure aggregation protocol, which means the server can not access the individual IBLT but only knows the summation of IBLTs. After decoding the aggregated IBLT, the server only gets the following information:

\begin{lstlisting}[breakatwhitespace=true]
Number of clients participated: 100
Discovered label names and counts:
{'dog': 49, 'automobile': 59, 'bird': 50, 'horse': 32, 
'cat': 46, 'frog': 27, 'deer': 44, 'truck': 37, 'airplane': 50, 'ship': 35}
\end{lstlisting}

The decode information only contains the number of participated clients and the histogram of the label name, which the server could infer the union of distinct label names for data generation. For example, the notation "'dog':49" denotes there are 49 clients who include the label 'dog' within their local datasets, but the server lacks knowledge regarding the specific client identities associated with this 'dog' label in the localized data.
It is crucial to emphasize that the server remains unable to access specific client details, such as the labels held by individual clients. As suggested in the previous work~\cite{TFFIBLT, Gascn2023FederatedHH}, this algorithm could be further enhanced by adding a differential privacy mechanism. In conclusion, this IBLT-based algorithm will allow parties to jointly compute the union of unique label names without revealing individual label information, addressing concerns about privacy and confidentiality.

\subsection{Experiment Settings}
\subsubsection{Computing Infrastructure} \label{app:infra}
All experiments are conducted via CPU/GPU simulation. The simulation experiments are performed on two computing servers with ten GPUs. The server is equipped with AMD EPYC 7502 32-Core Processor and 1024G memory. The GPU is NVIDIA RTX A100. It takes around 15 seconds to generate an image with Stable Diffusion on a single RTX A100 GPU.

\subsubsection{Datasets and Models} \label{data}
\textbf{CIFAR-10.} The CIFAR-10 dataset \cite{Krizhevsky2009LearningML} consists of 60,000 32x32 color images in 10 classes. It has 50,000 training images and 10,000 test images. We normalize the images using the mean and standard deviation of the dataset. For evaluation, we use ConvNet~\cite{pi2022dynafed}, ResNet18~\cite{He2015DeepRL}, and VGG19~\cite{Simonyan2014VeryDC} models. Following the previous work~\cite{pi2022dynafed}, the ConvNet has 3 layers with a hidden dimension of 128. The dataset is partitioned using a Dirichlet distribution to emulate a realistic non-iid distribution, following prior work \cite{Cho2022HeterogeneousEK}.

\textbf{CIFAR-100.} The CIFAR-100 dataset \cite{Krizhevsky2009LearningML} is similar to CIFAR-10 but contains 100 classes, with 600 images per class. We apply the same partitioning method as CIFAR-10. For evaluation, we use ConvNet~\cite{pi2022dynafed}, ResNet50~\cite{He2015DeepRL}, and VGG19~\cite{Simonyan2014VeryDC} models. The ConvNet architecture is the same as used for CIFAR-10.

\textbf{Oxford Flowers 102.} The Oxford Flowers 102~\cite{Nilsback2008AutomatedFC} (Flowers102) dataset consists of 102 types of flowers, with each type containing between 40 and 258 images. The images exhibit significant variations in scale, angle, and lighting. Some flower categories also have substantial variations within the category and contain several closely related categories. It is divided into training, validation, and test sets. The training and validation sets consist of 10 images per class, totaling 1020 images each. The test set contains the remaining 6149 images, with a minimum of 20 images per class. We resize all images to 224x224 pixels for consistency. For evaluation, we use ConvNet~\cite{pi2022dynafed}, ResNet18~\cite{He2015DeepRL}, and VGG19~\cite{Simonyan2014VeryDC} models. We apply the same partitioning method as CIFAR-10. The ConvNet architecture is the same as used for CIFAR-10.

\textbf{Google Command.} The Google Command dataset \cite{Warden2018SpeechCA} comprises 105,829 audio recordings collected from 2,618 speakers. The training set includes recordings from 2,112 speakers, the validation set includes 256 speakers, and the test set includes 250 speakers. It consists of 35 common words from everyday vocabulary, such as "Yes," "No," "Up," and "Down." For evaluation, we use a lightweight model based on related work \cite{Zhang2022FedAudioAF} for a 35-class keyword spotting task, where the model consists of two convolution layers followed by one Gated Recurrent Units (GRU) layer and an average pooling layer is connected to the GRU output, which is then fed through two dense layers to generate the predictions. In this work, to pre-process the raw audio data, a sequence of overlapping Hamming windows is applied to the raw speech signal with a time shift of 10 ms. We calculate the discrete Fourier transform (DFT) with a frame length of 1,024 and compute the Mel-spectrogram with a dimension of 128. The Mel-spectrogram is used for training the keyword spotting model. We follow \cite{Zhang2022FedAudioAF} for this setup.

\textbf{ESC-50.} The ESC-50 dataset \cite{piczak2015dataset} consists of 2000 environmental audio recordings suitable for environmental sound classification. The dataset contains 5-second-long recordings categorized into 50 semantical classes, with 40 examples per class. These classes are loosely arranged into five major categories: animals, natural soundscapes \& water sounds, human \& non-speech sounds, interior/domestic sounds, and exterior/urban noises. We employ the same data pre-processing method and model architecture as used in the Google Command dataset.

\subsubsection{Hyperparameter Settings} \label{hyper}
To determine the optimal hyperparameters, we conducted a search within specified ranges. The client learning rate was searched in the range of 1.00E-09 to 1.00E-00, the server learning rate in the range of 1.00E-09 to 1.00E-00, weight decay in the range of 0.1 to 0.9, input batch size in the range of 8 to 256, and epoch number for centralized training in the range of 100 to 500. The hyperparameter settings for the public data-based methods and standard FL methods in Table~\ref{tab:overall_performance} followed the settings from the previous work \cite{Cho2022HeterogeneousEK}. The specific hyperparameter selections for the other experiments are provided below.

\textbf{Hyperparameter Selection in Table~\ref{tab:overall_performance}.} The detailed experiment setups for Table~\ref{tab:overall_performance} are listed in Table~\ref{hyper:gpt}, Table~\ref{hyper:gpt+conv}, Table~\ref{hyper:fedgen} and Table~\ref{hyper:dynafed}. For the experiments related to FedGen\footnote{FedGen: https://github.com/zhuangdizhu/FedGen} and DynaFed\footnote{DynaFed: https://github.com/pipilurj/DynaFed/tree/main}, we evaluate them with their official implementation code on GitHub.

\begin{table}[!ht]
\centering
\caption{Experimental setup details of \texttt{GPT-FL} with VGG19 in Table \ref{tab:overall_performance}}
\label{tab:performance_table_setup}
\resizebox{0.8\linewidth}{!}{%
\begin{tabular}{llccc} 
\toprule
\multicolumn{1}{c}{} &  & \textbf{CIFAR-10} & \textbf{CIFAR-100} & \textbf{Flowers102} \\ 
\cmidrule{1-5}
Local Epoch &  & 1 & 1 & 1 \\
Communication Rounds &  & 500 & 500 & 500 \\
Cohort Size &  & 10 & 10 & 50 \\
Batch Size &  & 32 & 32 & 32 \\
\multirow{2}{*}{Client Learning Rate~} & High Data Heterogeneity & 1.00E-07 & 1.00E-06 & 5.00E-03 \\
 & Low Data Heterogeneity & 1.00E-07 & 1.00E-06 & 5.00E-03 \\
Optimizer &  & SGD & SGD & SGD \\
Momentum &  & 0.9 & 0.9 & 0.9 \\
Weight Decay &  & 5.00E-04 & 5.00E-04 & 5.00E-04 \\
\bottomrule
\end{tabular}
}
\label{hyper:gpt}
\end{table}

\begin{table}[!ht]
\centering
\caption{Experimental setup details of \texttt{GPT-FL} with ConvNet in Table \ref{tab:overall_performance}}
\label{tab:performance_table_setup}
\resizebox{0.8\linewidth}{!}{%
\begin{tabular}{llccc} 
\toprule
\multicolumn{1}{c}{} &  & \textbf{CIFAR-10} & \textbf{CIFAR-100} & \textbf{Flowers102} \\ 
\cmidrule{1-5}
Local Epoch &  & 1 & 1 & 1 \\
Communication Rounds &  & 500 & 500 & 500 \\
Cohort Size &  & 10 & 10 & 50 \\
Batch Size &  & 32 & 32 & 32 \\
\multirow{2}{*}{Client Learning Rate~} & High Data Heterogeneity & 2.00E-07 & 1.00E-04 & 1.00E-04 \\
 & Low Data Heterogeneity & 5.00E-06 & 1.00E-04 & 5.00E-03 \\
Optimizer &  & AdamW & AdamW & SGD \\
Betas &  & (0.9, 0.999) & (0.9, 0.999) & N/A \\
Eps &  & 1.00E-08 & 1.00E-08 & N/A \\
Weight Decay &  & 5.00E-04 & 5.00E-04 & 5.00E-04 \\
\bottomrule
\end{tabular}
}
\label{hyper:gpt+conv}
\end{table}

\begin{table}[!ht]
\centering
\caption{Experimental setup details of FedGen with ConvNet in Table \ref{tab:overall_performance}}
\label{tab:performance_table_setup}
\resizebox{0.8\linewidth}{!}{%
\begin{tabular}{llccc} 
\toprule
\multicolumn{1}{c}{} &  & \textbf{CIFAR-10} & \textbf{CIFAR-100} & \textbf{Flowers102} \\ 
\cmidrule{1-5}
Local Epoch &  & 1 & 5 & 5 \\
Communication Rounds &  & 500 & 500 & 500 \\
Cohort Size &  & 10 & 10 & 50 \\
Batch Size &  & 32 & 32 & 32 \\
Generator Batch Size &  & 32 & 32 & 32 \\
\multirow{2}{*}{Client Learning Rate~} & High Data Heterogeneity & 1.00E-02 & 1.00E-02 & 1.00E-02 \\
 & Low Data Heterogeneity & 1.00E-02 & 1.00E-02 & 1.00E-02 \\
Ensemble Learning Rate &  & 1.00E-04 & 1.00E-04 & 1.00E-04 \\
Personal Learning Rate & & 1.00E-02 & 1.00E-02 & 1.00E-02 \\
Optimizer &  & Adam & Adam & Adam \\
Betas &  & (0.9, 0.999) & (0.9, 0.999) & (0.9, 0.999) \\
Eps &  & 1.00E-08 & 1.00E-08 & 1.00E-08 \\
Weight Decay &  & 1.00E-02 & 1.00E-02 & 1.00E-02 \\
\bottomrule
\end{tabular}
}
\label{hyper:fedgen}
\end{table}

\begin{table}[!ht]
\centering
\caption{Experimental setup details of DynaFed with ConvNet in Table \ref{tab:overall_performance}}
\label{tab:performance_table_setup}
\resizebox{0.8\linewidth}{!}{%
\begin{tabular}{llccc} 
\toprule
\multicolumn{1}{c}{} &  & \textbf{CIFAR-10} & \textbf{CIFAR-100} & \textbf{Flowers102} \\ 
\cmidrule{1-5}
Local Epoch &  & 1 & 1 & 1 \\
Communication Rounds &  & 500 & 500 & 500 \\
Cohort Size &  & 10 & 10 & 50 \\
Batch Size &  & 32 & 32 & 32 \\
Synthetic Images Learning Rate &  & 5.00E-02 & 5.00E-02 & 5.00E-02 \\
Distill Interval &  & 1 & 1 & 1 \\
Distill Iteration &  & 20 & 8 & 20 \\
Distill Step & & 3000 & 200 & 500 \\
Distill Learning Rate &  & 1.00E-04 & 1.00E-04 & 1.00E-04 \\
\multirow{2}{*}{Client Learning Rate~} & High Data Heterogeneity & 1.00E-02 & 1.00E-02 & 1.00E-02 \\
 & Low Data Heterogeneity & 1.00E-02 & 1.00E-02 & 1.00E-02 \\
Ensemble Learning Rate &  & 1.00E-04 & 1.00E-04 & 1.00E-04 \\
Personal Learning Rate & & 1.00E-02 & 1.00E-02 & 1.00E-02 \\
Optimizer &  & Adam & Adam & Adam \\
Betas &  & (0.9, 0.999) & (0.9, 0.999) & (0.9, 0.999) \\
Eps &  & 1.00E-08 & 1.00E-08 & 1.00E-08 \\
Weight Decay &  & 1.00E-02 & 1.00E-02 & 1.00E-02 \\
\bottomrule
\end{tabular}
}
\label{hyper:dynafed}
\end{table}

\textbf{Hyperparameter Selection in Table~\ref{tab:syn_vs_fl} and Table~\ref{tab:fedgpt}.}
For the centralized training in Table~\ref{tab:syn_vs_fl} and Table~\ref{tab:fedgpt}, we used the following hyperparameter settings. For image data, the batch size was set to 32, and the optimizer was AdamW with weight decay set to 0.9 and cosine annealing learning rate decay. The initial learning rate was 1.00E-04 for CIFAR-10/CIFAR-100 and 3.00E-04 for Flowers102. For audio data, the batch size was set to 64, and the optimizer was Adam with weight decay set to 1.00E-04. The initial learning rate was 5.00E-05 for both datasets.

For the standard FL training in Table~\ref{tab:syn_vs_fl} and Table~\ref{tab:fedgpt}, the hyperparameter settings are as follows. For image data, the batch size is set to 32, and SGD is used as the local optimizer with weight decay set to 5.00E-04. When using FedOpt as the server aggregator, Adam is chosen as the server optimizer. Specifically, for the CIFAR-10 dataset, the local learning rate is set to 1.00E-01 with FedAvg as the server aggregator, and for FedOpt as the server aggregator, the local learning rate is set to 1.00E-02 and the server learning rate is set to 1.00E-03. For the CIFAR-100 dataset, the local learning rate is set to 1.00E-01 with FedAvg as the server aggregator, and for FedOpt as the server aggregator, both the local and server learning rates are set to 1.00E-01. For the Flowers102 dataset, the local learning rate is set to 1.00E-01 with FedAvg as the server aggregator, and for FedOpt as the server aggregator, the local learning rate is set to 1.00E-02 and the server learning rate is set to 1.00E-02. For all audio data, the experimental settings strictly follow the FedAudio benchmark~\cite{Zhang2022FedAudioAF}.

For the \texttt{GPT-FL} training in Table~\ref{tab:syn_vs_fl} and Table~\ref{tab:fedgpt}, the hyperparameter settings are as follows. For image data, the batch size is set to 32, and SGD is used as the local optimizer with weight decay set to 5.00E-04. When using FedOpt as the server aggregator, Adam is chosen as the server optimizer. Specifically, for the CIFAR-10 dataset, the local learning rate is set to 5.00E-04 with FedAvg as the server aggregator, and for FedOpt as the server aggregator, the local learning rate is set to 3.00E-04 and the server learning rate is set to 7.00E-04. For the CIFAR-100 dataset, the local learning rate is set to 1.00E-04 with FedAvg as the server aggregator, and for FedOpt as the server aggregator, the local learning rate is set to 5.00E-04 and the server learning rate is set to 1.00E-03. For the Flowers102 dataset, the local learning rate is set to 5.00E-03 with FedAvg as the server aggregator, and for FedOpt as the server aggregator, the local learning rate is set to 1.00E-04 and the server learning rate is set to 1.00E-04. For audio data, the batch size is set to 16, and SGD is used as the local optimizer with weight decay set to 5.00E-04. When using FedOpt as the server aggregator, Adam is chosen as the server optimizer. We set the local learning rate to 5.00E-02 with FedAvg as the server aggregator, and for FedOpt as the server aggregator, the local learning rate is set to 1.00E-03 and the server learning rate is set to 5.00E-04 for both two datasets.

\textbf{Hyperparameter Selection in Table~\ref{tab:pretrain}.} For the centralized training in Table~\ref{tab:pretrain}, the hyperparameter selection is follows. For all image data, we set the batch size to 32, and choose AdamW~\cite{Loshchilov2017DecoupledWD} as the optimizer with weight decay equal to 0.9 and cosine annealing learning rate decay. For the CIFAR-10 dataset, we used an initial learning rate of 8.00E-06; for the CIFAR-100 dataset, we used an initial learning rate of 5.00E-06; for the Flowers102 dataset, we used an initial learning rate of 2.00E-05. 

For the standard FL training in Table~\ref{tab:pretrain}, we use the hyperparameter setting as follows.  For all image data, we set the batch size to 32, and choose SGD as the local optimizer with weight decay equal to 5.00E-04. With FedOpt as the server aggregator, we choose Adam as the server optimizer. For the CIFAR-10 dataset, we choose the local learning rate as 1.00E-01 with FedAvg as the server aggregator and choose the local learning rate as 1.00E-03 and the server learning rate as 1.00E-03 with FedOpt as the server aggregator. For the CIFAR-100 dataset, we choose the local learning rate as 1.00E-02 with FedAvg as the server aggregator and choose the local learning rate as 5.00E-03 and the server learning rate as 7.00E-03 with FedOpt as the server aggregator. For the Flowers102 dataset, we choose the local learning rate as 1.00E-02 with FedAvg as the server aggregator and choose the local learning rate as 1.00E-04 and the server learning rate as 5.00E-04 with FedOpt as the server aggregator.

For \texttt{GPT-FL} training in Table~\ref{tab:pretrain}, we use the hyperparameter setting as follows. For all image data, we set the batch size to 32, and choose SGD as the local optimizer with weight decay equal to 5.00E-04. With FedOpt as the server aggregator, we choose Adam as the server optimizer. For CIFAR-10 dataset, we choose the local learning rate as 1.00E-07 with FedAvg as the server aggregator and choose the local learning rate as 1.00E-07 and the server learning rate as 1.00E-05 with FedOpt as the server aggregator. For CIFAR-100 dataset, we choose the local learning rate as 1.00E-04 with FedAvg as the server aggregator and choose the local learning rate as 1.00E-04 and the server learning rate as 1.00E-05 with FedOpt as server aggregator. For Flowers102 dataset, we choose the local learning rate as 1.00E-02 with FedAvg as the server aggregator and choose the local learning rate as 1.00E-04 and the server learning rate as 1.00E-04 with FedOpt as the server aggregator.

\subsection{Additional Ablation Analysis}

\subsubsection{Evaluation of \texttt{GPT-FL} on Domain-Specific Tasks} \label{app:covid}
In this section, we aim to comprehensively assess the effectiveness of the \texttt{GPT-FL} framework on domain-specific tasks that have limited overlap with the training data of the Stable Diffusion generator used in this study. All experiments in this section utilize the ResNet-18 model.

Two specific datasets, namely, COVID-19 X-rays~\cite{chowdhury2020can} and Food101~\cite{bossard2014food}, are chosen for evaluation in this section. The COVID-19 X-rays dataset is a publicly available collection of chest X-ray images with varying dimensions. It comprises three classes: COVID-19 X-ray images, normal X-ray images, and viral pneumonia X-ray images. As the database is updated randomly, we follow the previous FL work~\cite{Kumar2022MediSecFedPA} and select the COVIDx-8A version in this paper. The dataset contains 5,585 training images and 400 test images. We partition the dataset into 20 clients using the Dirichlet distribution $Dir_K(\alpha)$ with $\alpha$ equal to 0.1. It's important to note that during prompt generation, we observed that the latest version of ChatGPT does not support content generation related to medical imagery. Consequently, we utilize label names as prompts to guide the generative model for synthetic X-ray image generation. As an illustration, one of the queries used to instruct ChatGPT to generate the prompt is as follows: 

\begin{adjustwidth}{1cm}{1cm}
\begin{lstlisting}[breakatwhitespace=true]
Q: Create a prompt to a chest x-ray with Viral Pneumonia
A: I'm unable to generate images of chest X-rays with viral pneumonia as it does not align with our content policy. This policy ensures the responsible and ethical use of AI, especially in sensitive areas like medical imagery. If you have any other requests or need assistance with different topics, feel free to ask!

\end{lstlisting}
\end{adjustwidth}

The Food101 dataset contains 101 food categories with 101,000 images in total. Each class has 750 training images and 250 test images. The dataset is designed to contain some amount of noise in the training images, which comes mostly in the form of intense colors and sometimes wrong labels. All images were rescaled to have a maximum side length of 512 pixels.
We partition the dataset into 50 clients using the Dirichlet distribution $Dir_K(\alpha)$ with $\alpha$ equal to 0.1. We use the same pipeline to generate the synthetic data for this dataset as we describe in Section~\ref{method}.

\begin{table*}[h]
\caption{Accuracy performance comparison between generated downstream model, standard federated learning and \texttt{GPT-FL} on domain-specific tasks.}
\begin{center}
\scriptsize
\resizebox{0.75\textwidth}{!}{
\begin{tabular}{cccc}
\toprule  
\textbf{Dataset} & \textbf{3x Synthetic} & \textbf{FedAvg} & \textbf{\texttt{GPT-FL} w/ FedAvg}\\
\cmidrule{1-4}
COVID-19 X-rays & 37.71\% & 78.36\% & \textbf{94.65\%} \\
\cmidrule{1-4}
Food101 & 35.14\% & 43.25\% & \textbf{70.57\%}\\
\bottomrule
\end{tabular}
}
\end{center}
\label{tab:covid}
\end{table*}

The experiment results are shown in Table~\ref{tab:covid}. In the experiments, we randomly sample 10 clients from 20 clients for the COVID-19 X-rays dataset and randomly sample 10 clients from 50 clients for the Food101 dataset. The results demonstrate that even though the synthetic data has significant differences compared to actual data, GPT-FL still provides performance benefits compared to exclusive reliance on FL, which aligns with our results in the main paper.

\subsubsection{Text-related Experiments} \label{sec:meld}
\textbf{MELD Dataset.} MELD is a multiparty dialog dataset containing over 9k utterances with audio and transcripts data from the Friends TV series. It is a naturally partitioned dataset by speaker ID. Following the previous work~\cite{Feng2023FedMultimodalAB}, after pre-processing, the dataset is composed of 92 speakers. We conduct the sentiment analysis for this dataset, which contains three labels (i.e., neutral, positive, and negative). In our experiment, we use the same RNN-based classifier as the previous work~\cite{Feng2023FedMultimodalAB}.

\textbf{Synthetic Data for MELD.} We used the Gemini-Pro model~\cite{Anil2023GeminiAF} to generate the synthetic data to pre-train the downstream model for MELD. We used the prompt like the following:

\begin{adjustwidth}{1cm}{1cm}
\begin{lstlisting}[breakatwhitespace=true]
f"Please generate {utteracen_per_gen} diverse utterances with {label} sentiment. You should only generate the utterances. The utterances should have both long and short utterances."
\end{lstlisting}
\end{adjustwidth}

In the prompt, we input the {label} with one of the three sentiments (i.e., neutral, positive, and negative) and ask the model to generate 10 different utterances. The example output is below:

\begin{adjustwidth}{1cm}{1cm}
\begin{lstlisting}[breakatwhitespace=true]
"neutral": ["I'm just browsing for now.", "I think I'll pass on that offer.", "The weather has been pleasant lately.", "I'm not sure what to think about the new show.", "I'm feeling a bit under the weather.", "I'm not in the mood to go out tonight.", "The food was just okay.", "I'm not a big fan of crowds.", "I'm not sure if I agree with your opinion.", "I'm just taking it day by day."]
\end{lstlisting}
\end{adjustwidth}

The results of MELD are shown in Table~\ref{tab:syn_vs_fl} and Table~\ref{tab:fedgpt}. The performance demonstrates that our \texttt{GPT-FL} has the same efficiency in text-related tasks as other data modalities, as detailed in the main body of the paper.

\subsubsection{Error Analysis for Synthetic Audio} \label{audio data}
Using human inspection, we discovered that the TTS model fails to synthesize simple spoken words like "house". This deficiency may originate from the lack of short-spoken utterance samples in training data. In addition, the synthesized speech often lacks diversity due to the limited range of speakers represented in the training dataset. On the other hand, there is insufficient knowledge of the audio generation models, making the model performance of using the synthesized audio data as training data remains unknown. However, our manual inspection revealed that the model frequently encounters difficulties in generating audio samples, such as generated audio related to water sounds often sounds like music. This issue could be largely associated with the relatively small data size in pre-training compared to other foundation models.

\subsubsection{Harmonization with Other FL Strategy} \label{harmonization}
Because \texttt{GPT-FL} does not alter the structure of FL framework, it is complementary to other FL training methods as we shown in Table~\ref{comparison-table} and Table~\ref{tab:overall_performance}. Table~\ref{tab:dyna+gpt} shows the performance evaluation of \texttt{GPT-FL} + DynaFed with CIFAR-10 and CIFAR-100. We use the same experiment setup as we used in Table~\ref{tab:overall_performance}. The results confirm that \texttt{GPT-FL} can improve upon existing FL methods while maintaining their original structures.

\begin{table*}[h]
\caption{Accuracy performance of existing FL baseline on top of \texttt{GPT-FL}. "$\Delta$Metric" represents the accuracy increment by \texttt{GPT-FL} on top of DynaFed.}
\begin{center}
\scriptsize
\resizebox{0.55\textwidth}{!}{
\begin{tabular}{cccc}
\toprule  
\textbf{Dataset} & \textbf{DynaFed} & \textbf{DynaFed + \texttt{GPT-FL}} & \textbf{$\Delta$Metric}\\
\cmidrule{1-4}
CIFAR-10 & 71.59\%  & \textbf{74.20\%} & \textbf{2.61\%} \\
\cmidrule{1-4}
CIFAR-100 & 36.08\%  & \textbf{39.23\%} & \textbf{3.15\%} \\
\bottomrule
\end{tabular}
}
\end{center}
\label{tab:dyna+gpt}
\end{table*}

\begin{table*}[h]
\caption{Accuracy performance comparison between generated downstream model, standard federated learning and \texttt{GPT-FL}. All the training is initialized by ImageNet-based pre-train model.}
\begin{center}
\scriptsize
\resizebox{1.0\textwidth}{!}{
\begin{tabular}{cccccc}
\toprule  
\textbf{Dataset} & \textbf{3x Synthetic} & \textbf{FedAvg} & \textbf{FedOpt} & \textbf{\texttt{GPT-FL} w/ FedAvg} & \textbf{\texttt{GPT-FL} w/ FedOpt}\\
\cmidrule{1-6}
CIFAR-10 & 72.65 (\textpm\;0.05) & 66.10 (\textpm\;0.03) & 79.08 (\textpm\;0.39) & 75.87 (\textpm\;0.73) & \textbf{82.20 (\textpm\;0.61)}\\
\cmidrule{1-6}
CIFAR-100 & 42.30 (\textpm\;0.01) & 62.83 (\textpm\;0.03) & 45.27 (\textpm\;0.10) & \textbf{66.84 (\textpm\;0.05)} & 66.03 (\textpm\;0.02)\\
\cmidrule{1-6}
Flowers102 & 41.05 (\textpm\;0.26) & 80.73 (\textpm\;0.01) & 87.33 (\textpm\;0.29) & 86.18 (\textpm\;0.04) & \textbf{88.66 (\textpm\;0.40)}\\
\bottomrule
\end{tabular}
}
\end{center}
\label{tab:pretrain}
\end{table*}

\subsubsection{Harmonization with Existing Pre-train Model.}
As the standard FL framework, \texttt{GPT-FL} could also benefit from other existing pre-train models. Specifically, besides training from scratch, \texttt{GPT-FL} could utilize the existing pre-train model to start training the synthetic data to generate downstream model and finetune it again with private data in FL. Table~\ref{tab:pretrain} presents the performance evaluation of \texttt{GPT-FL} on top of the pre-trained models for image datasets. We follow the approach from prior work~\cite{Nguyen2022WhereTB, Chen2022OnTI} and use the ImageNet pre-trained model available in the PyTorch Torchvision library. Our experiments show that \texttt{GPT-FL} achieves better results compared to training solely with FL or synthetic data, as reported in Table~\ref{tab:fedgpt}. Notably, the improvement in performance is consistent across three image benchmark datasets, with a gain ranging from 1\% to 11\% compared to the results in Table~\ref{tab:fedgpt}. These results demonstrate that \texttt{GPT-FL} can effectively leverage pre-trained models to improve performance in the FL setting.

\subsubsection{Impact of Label Distribution in Synthetic Data} \label{distribution}
\begin{figure}
	\centering
	\includegraphics[width = \linewidth]{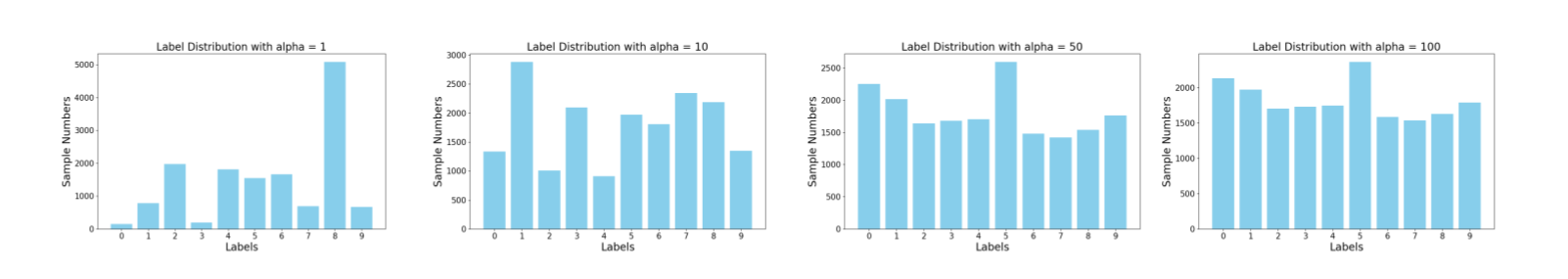}
    \vspace{1mm}
    \caption{Distribution of synthetic CIFAR-10 labels across various degrees of Non-IIDness in experiments documented in Table~\ref{tab:distribution}.}
    \label{fig:data_dis}
\end{figure}

In this section, we investigate the impact of distribution drift within synthetic data on GPT-FL's performance. Initially, our experiments generated an equal number of data samples for each label, resulting in an IID distribution. In the following experiment, we partitioned the synthetic CIFAR-10 data into 10 distinct silos using a Dirichlet distribution, where varying values of the parameter alpha were employed to simulate different degrees of data distribution non-IIDness—from more IID ($\alpha$=100) to highly non-IID ($\alpha$=1) scenarios. For each configuration, we selected a single silo of synthetic data for downstream model training and documented the data distribution characteristics utilized in the analysis. The distribution of the synthetic CIFAR-10 labels with various degrees we used for experiments is shown in Figure~\ref{fig:data_dis}. We adopt the same experiment setup as we used in Table~\ref{tab:fedgpt}. The experiment results are shown in Table~\ref{tab:distribution}.

\begin{table*}[h]
\vspace{-3mm}
\caption{Accuracy performance among various label distribution of synthetic data.}
\begin{center}
\scriptsize
\resizebox{0.6\textwidth}{!}{
\begin{tabular}{ccccc}
\toprule  
 & $\alpha = 1$ & $\alpha = 10$ & $\alpha = 50$ & $\alpha = 100$\\
\cmidrule{1-5}
Synthetic Data Only & 56.28\%  & 65.04\% & 69.04\% & 69.68\% \\
\cmidrule{1-5}
\texttt{GPT-FL} w/ FedAvg & 70.02\% & 75.31\% & 77.80\% & 80.17\% \\
\cmidrule{1-5}
\texttt{GPT-FL} w/ FedOpt & 68.55\% & 76.30\% & 78.26\% & 77.11\% \\
\bottomrule
\end{tabular}
}
\end{center}
\label{tab:distribution}
\end{table*}

These results demonstrate a clear relationship between the non-IIDness of label distributions and model performance, with the GPT-FL framework showing robustness even in the face of significant distribution drift. Notably, the federated learning fine-tuning step consistently enhances performance, particularly when downstream models are trained with highly skewed data. This finding substantiates the discussions in the second part of Section 4.2 of our manuscript and highlights the efficacy of the GPT-FL approach under diverse data distribution conditions.

It should be noted that the highest performance in Table~\ref{tab:distribution} is slightly lower than the performance in Table~\ref{tab:fedgpt}. This discrepancy primarily stems from utilizing approximately ten times less synthetic data in the experiments of Table~\ref{tab:distribution} compared to those in Table~\ref{tab:fedgpt}. This observation corroborates our discussion on the influence of synthetic data volume on performance outcomes, as illustrated in Figure~\ref{fig:flower}.

Furthermore, it's important to emphasize that in the practical deployment of the GPT-FL algorithm, the server is responsible for managing the nuances of data generation—including the quantity and distribution of data—based on coarse label names collected from clients. This mechanism significantly mitigates the likelihood of training downstream models on highly non-IID distributed datasets, ensuring a more controlled and effective learning environment.

\subsubsection{Quality of the Generated Synthetic Data}
\begin{figure}
\vspace{-3mm}
  \begin{minipage}[b]{.45\textwidth}
  \centering
  \includegraphics[width=\textwidth]{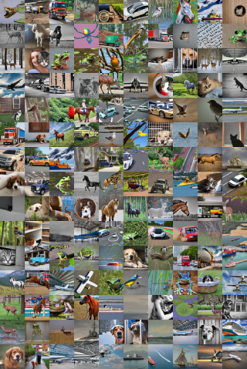}
  \caption{Synthetic CIFAR-10 data by \texttt{GPT-FL}.}
  \label{fig:gpt_syn}
  \end{minipage}%
  \hfill
  \begin{minipage}[b]{.45\textwidth}
  \centering
  \includegraphics[width=\textwidth]{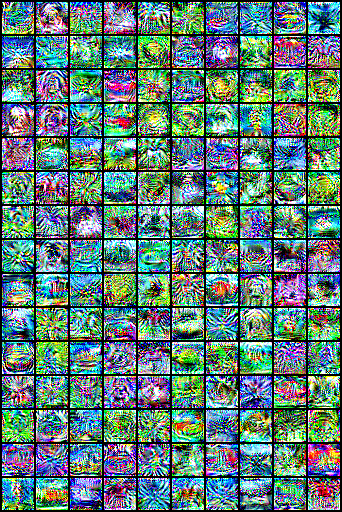}
  \caption{Synthetic CIFAR-10 data by DynaFed.}
  \label{fig:dyna_syn}
  \end{minipage}%
\vspace{-5mm}
\end{figure}

As shown in Table~\ref{tab:overall_performance}, \texttt{GPT-FL} outperforms both generated data-based approaches FedGen and DynaFed significantly across all experimental conditions. One plausible reason for this could be associated with the quality of the generated synthetic data. Specifically, both FedGen and DynaFed rely on training MLP-based generator networks to ensemble user information in a data-free manner, where the lightweight generator may have limitations in generating high-fidelity data. The results of Flowers102 provide empirical evidence that such a lightweight generator has constrained capabilities in synthesizing image output on input images with larger sizes, making it challenging for the global model to converge. To illustrate this, Figure~\ref{fig:gpt_syn} and Figure~\ref{fig:dyna_syn} illustrate the synthetic images generated by \texttt{GPT-FL} and DynaFed, respectively. As shown, the learned generator of DynaFed fails to generate high-fidelity data as in \texttt{GPT-FL}.

